%% file: acl_latex.tex
\pdfoutput=1

\documentclass[11pt]{article}
\usepackage{bm}
 
\usepackage[]{acl}

\usepackage{times}
\usepackage{latexsym}

\usepackage[T1]{fontenc}

\usepackage[utf8]{inputenc}

\usepackage{microtype}

\usepackage{inconsolata}

\usepackage{pgfplots}
\pgfplotsset{compat=1.18}
\usepackage[edges]{forest}

\usepackage{graphicx}
\input{commands.tex}

\title{A Survey on Evaluation of LLM-based Agents}

\author{
  \textbf{Asaf Yehudai\textsuperscript{1,2}},
  \textbf{Lilach Eden\textsuperscript{2}},
  \textbf{Alan Li\textsuperscript{3}},
  \textbf{Guy Uziel\textsuperscript{2}},
\\
  \textbf{Yilun Zhao\textsuperscript{3}},
  \textbf{Roy Bar-Haim\textsuperscript{2}},
  \textbf{Arman Cohan\textsuperscript{3}},
  \textbf{Michal Shmueli-Scheuer\textsuperscript{2}}
\\
\\
  \textsuperscript{1}The Hebrew University of Jerusalem
  \textsuperscript{2}IBM Research
  \textsuperscript{3}Yale University
\\
\{Asaf.Yehudai, Guy.Uziel1\}@ibm.com  
\{lilache, roybar, shmueli\}@il.ibm.com \\
\{haoxin.li, yilun.zhao, arman.cohan\}@yale.edu
}

\begin{document}
\maketitle

\begin{abstract}
LLM-based agents represent a paradigm shift in AI, enabling autonomous systems to plan, reason, and use tools while interacting with dynamic environments. This paper provides the first comprehensive survey of evaluation methods for these increasingly capable agents. We analyze the field of agent evaluation across five perspectives: (1) Core LLM capabilities needed for agentic workflows, like planning, and tool use; (2) Application-specific benchmarks such as web and SWE agents; (3) Evaluation of generalist agents; (4) Analysis of agent benchmarks' core dimensions; and (5) Evaluation frameworks and tools for agent developers. Our analysis reveals current trends, including a shift toward more realistic, challenging evaluations with continuously updated benchmarks. We also identify critical gaps that future research must address—particularly in assessing cost-efficiency, safety, and robustness, and in developing fine-grained, scalable evaluation methods \footnote{Our \href{https://github.com/Asaf-Yehudai/LLM-Agent-Evaluation-Survey}{GitHub repository} tracks works in the field.}.

\end{abstract}

\section{Introduction}

LLMs have recently made remarkable progress, tackling a wide range of challenging tasks. Yet, LLMs are static, having fixed knowledge, and confined to text-to-text interaction. \texttt{LLM-based agents} address those gaps by building on LLMs as a backbone, integrating them into multi-step workflows and equipping them with external tools~\cite{wang2024survey}. Hence, LLM agents can perform computations, retrieve up-to-date information, and interact with their environment. Crucially, they can autonomously plan, execute, and adapt complex strategies in real-world settings. This agency enables them to tackle problems once beyond the reach of AI, unlocking innovative applications across diverse domains.


\input{figures/layout_fig_short}

The shift from static models to adaptive, interactive agents calls for a new paradigm for \textit{evaluating} LLM-based agents.
Such evaluation must go beyond measuring LLM textual outputs to assess an agent’s capacity for sequential decision-making and operation within dynamic environments. 
It requires benchmarks that can assess the agent's ability to accomplish user tasks via a sequence of actions and interactions.
Moreover, benchmarks must co-evolve with agent capabilities, accommodating new classes of tasks and domains.

In this survey, we present the first overview of LLM-based agent evaluation. We aim to benefit developers, benchmark creators, practitioners, and researchers by mapping the current evaluation landscape and identifying key gaps for future research.

We begin by discussing the evaluation of fundamental LLM-based agent capabilities (\S\ref{sec:capabilities}). These include planning, tool use, self-reflection, and memory. We then review benchmarks and evaluation strategies for prominent types of agentic applications: web agents, software engineering agents, scientific agents, and conversational agents (\S\ref{sec:use_case}).  
We continue to describe benchmarks and leaderboards for evaluating generalist agents (\S\ref{sec:general}). Consequently, we define and analyze core dimensions of agent benchmarks (\S\ref{sec:bench_dims}). \S\ref{sec:frameworks} reviews current evaluation frameworks for agent developers. These frameworks integrate with the agent's development environment, and support its evaluation throughout the entire development cycle. We conclude with a discussion (\S\ref{sec:dis}) of current trends and future research directions in agent evaluation. Figure~\ref{fig:agent-evaluation-typology} offers a visual summary of the survey’s structure. 





\section{Agent Capabilities Evaluation}
\label{sec:capabilities}

LLM-based agents are composed of a backbone LLM and an agent harness~\cite{Yao2022ReActSR}. Thus, evaluating the core suite of LLM abilities required for agentic tasks is paramount to understanding the potential and limitations of LLM-based agents.
Each ability can be evaluated in isolation or as part of a full agent workflow. 
Here, we shortly describe four such core agent abilities. In Appendix \S\ref{app:capabilities} we provide a more detailed review of each one.


\paragraph{Planning and Multi-Step Reasoning}\label{sec:plan}
enables agents to decompose problems into smaller, more manageable subtasks and create strategic execution paths toward solutions \cite{gao2023largelanguagemodelsempowered}.

LLM reasoning benchmarks requiring multiple logical steps, such as HotpotQA~\cite{yang2018hotpotqadatasetdiverseexplainable, cobbe2021gsm8k,suzgun2022bbh}, have been used to evaluate agent-based approaches like ReAct. 
More specialized planning benchmarks, such as PlanBench~\cite{valmeekam2023planbench}, adapt classical planning tasks to assess LLMs and reveal gaps in long-term planning~\cite{stein2023autoplanbench}.
Agent-oriented planning benchmarks further evaluate an agent’s ability to follow structured workflows~\cite{xiao2024flowbench} or to manage real-world planning tasks expressed in natural language~\cite{zheng2024natural}, with focus on long-horizon planning with verifiable constraints~\cite{zhang2026deepplanningbenchmarkinglonghorizonagentic}. 
Results show that even SOTA models struggle with long-horizon planning.

\paragraph{Function Calling \& Tool Use}\label{sec:tool}
is a fundamental ability for agents to deliver updated, contextually accurate responses \cite{qin2023toolllm, tang2023toolalpaca}. 
Function calling involves several subtasks that work together seamlessly, including intent recognition, function selection, and parameter-value pair-mapping.



Initial benchmarks for tool use focused on these sub-tasks, providing relatively simple, one-step interactions with explicitly predefined parameters.  Benchmarks such as ToolAlpaca \cite{tang2023toolalpaca}, ToolBench \cite{qin2023toolllm}, and the Berkeley Function Calling Leaderboard v1 (BFCL) \cite{berkeley-function-calling-leaderboard} represent this early stage, relying on synthetic data and rule-based matching to measure metrics like pass rates and structural accuracy.


Later versions of BFCL (v2 and v3) introduced multi-turn interactions, organizational tools, and multi-step logic, emphasizing continuous state management. Furthermore, NESTFUL~\cite{basu2024nestful} introduces cases where calls are dependent on previous ones, while ComplexFuncBench~\cite{zhong2025complexfuncbench} presents scenarios requiring implicit parameter inference, adherence to user-defined constraints, and efficient long-context processing. 


To better reflect real-world scenarios, recent evaluations have become increasingly agentic~\cite{patil2025bfcl}, requiring longer interactions and scaling the number of domains and tools by sourcing them from real MCP servers~\cite{anthropic2024mcp}. Two current frontier benchmarks, Scale's MCP Atlas~\cite{bandi2026mcpatlaslargescalebenchmarktooluse} and Tool-Decathlon~\cite{li2026the}, further push in this direction. Despite significant model advancements, these benchmarks continue to pose challenges.

\paragraph{Self-Reflection}\label{sec:self_reflect}
enables agents to self-correct by
dynamically adjusting reasoning or actions based on feedback~\cite{Renze_2024}.


Early evaluation efforts repurposed existing benchmarks into multi-turn feedback loops to gauge LLM's abilities to self-correct~\citep{renze2024selfreflectionllmagentseffects,huang2024largelanguagemodelsselfcorrect,Shinn2023ReflexionLA,you-etal-2024-llm}. In agentic settings, works like LLM-Evolve~\citep{you-etal-2024-llm} reuse past feedback as in-context examples to evaluate self-reflection. Similarly, LLF-Bench~\citep{cheng2023llfbenchbenchmarkinteractivelearning} utilizes feedback to assess decision-making in diverse environments.
Despite these efforts, a standardized benchmark or methodology for assessing self-reflection remains a critical gap.

\paragraph{Memory}\label{sec:memory} mechanisms enable LLM agents to manage information and reason across extended interactions~\cite{park2023generative}, supporting different memory types: episodic (past interactions), semantic (factual knowledge), and procedural (operational information)~\cite{hatalis2023memory}.


Early studies employed long-context benchmarks~\cite{liu2024lost, pang2021quality} to assess memory mechanisms~\cite{packer2024memgptllmsoperatingsystems, xu2025amemagenticmemoryllm}.
More recently, dedicated benchmarks for agentic memory have been introduced.
For episodic memory, Benchmarks evaluate how agents leverage prior interactions and feedback to support continual improvement across multi-session agentic tasks~\cite{wu2024streambench, he2026memoryarenabenchmarkingagentmemory}.
For semantic memory, benchmarks assess retrieval effectiveness and long-range understanding, revealing that current methods remain limited in maintaining long-range consistency and handling dynamic memory~\cite{tan-etal-2025-membench, hu2025evaluating, wu2025longmemevalbenchmarkingchatassistants}.

\section{Application-Specific Agents Evaluation} \label{sec:use_case}

The landscape of application-specific agents is expanding, with an increasing number of specialized agents emerging across tasks and domains such as web, software, game, embodied, search, and scientific agents~\cite{wang2024survey}. Here, we focus on four representative and prominent applications.
We review these agents while implicitly addressing several core benchmark dimensions, such as data curation, environment type, and metrics, which we discuss systematically in \S\ref{sec:bench_dims}.


\subsection{Web Agents}\label{sec:web}



Web agents handle web-related tasks, including e-commerce, information search, and personal assistant tasks.
Early work presented simplified simulation environments with limited interaction options~\citep{pmlr-v70-shi17a, liu2018reinforcementlearningwebinterfaces}. For example, WebShop~\citep{yao2022webshop} simulates online shopping tasks, from product search to checkout.



More recently, the field shifted toward more realistic evaluation environments.
The two most widely used offline and sandboxed online environments are Mind2Web~\citep{deng2023mind2web} and WebArena~\citep{zhou2023webarena}, respectively. These environments serve as the foundation for most current evaluation efforts.
Mind2Web provides an offline environment with real websites across diverse domains, supporting rich user interactions (e.g., clicking, selecting, or typing into any element) and enabling intermediate goal evaluation by comparing predicted operations with gold-standard actions.
In contrast, WebArena introduces a dynamic environment featuring fully functional websites across multiple domains, enriched with auxiliary tools and knowledge sources.
It also defines diverse, long-horizon human tasks, with corresponding functional correctness tests.

Subsequent works build on these environments to evaluate specific dimensions of web interaction, focusing on multi-turn dialogue~\citep{lu2024weblinx}, office and enterprise workflows performed by knowledge workers~\citep{drouin2024workarena, boisvert2025workarena++}, and multi-site, time-intensive tasks~\citep{yoran2024assistantbench}.
Other works refine the evaluation process, offering a more granular analysis of agent performance~\citep{pan2024webcanvas}, or employing LLM-as-a-judge methods for more semantic, human-aligned evaluation~\citep{xue2025illusion}.
Importantly, \citet{levy2024st} emphasizes safety and trustworthiness by assessing policy compliance and risk mitigation, which are important to real-world deployment.

An important line of research emphasizes the multimodal nature of the modern web, with benchmarks that require agents to integrate visual and textual information, typically interacting via a graphical interface~\citep{koh2024visualwebarena, zhang2024mmina}.
Notably, WebVoyager~\citep{he2024webvoyager} has gained significant commercial interest due to its realistic multimodal online evaluation. However, recent work suggests it exhibits over-optimistic performance estimates, and proposes Online-Mind2Web as a more rigorous alternative that remains challenging for current agents~\citep{xue2025illusion}.

\subsection{Software Engineering Agents} \label{sec:se}

The evaluation of software engineering (SWE) agents began with benchmarks that measured LLM fundamental coding capabilities \cite{Chen2021EvaluatingLL, Austin2021ProgramSW}. These early benchmarks focused on short, self-contained, algorithm-specific tasks, thus falling short of addressing the full complexity of real-world SWE tasks.



SWE-bench~\cite{Jimenez2023SWEbenchCL} was introduced to address the above shortcomings by providing an end-to-end evaluation framework grounded in real-world GitHub issues. Each task includes a detailed issue description, full repository context, executable environments (e.g., Docker), and validation tests, enabling agents to generate and verify code patches automatically. 
Several follow-up works identified evaluation issues, like overly specific or unrelated unit tests, underspecified issue descriptions, and problematic environmental setups, and thus proposed new variants~\cite{swebenchlite2024, Xia2024AgentlessDL, swebenchverified2024, Aleithan2024SWEBenchEC}.
The most widely adopted of these, SWE-bench Verified~\cite{swebenchverified2024}, carries extensive human filtration and validation to yield a high-quality subset of 500 samples.
It also standardizes execution through containerized environments and provides difficulty annotations, enabling more reproducible and interpretable evaluations. These improvements make SWE-bench Verified the de facto benchmark for assessing SWE agents.



Following the success of SWE-bench, multilingual~\cite{zan2024swe, yang2025swesmith} and multimodal~\cite{Yang2024SWEbenchMD} versions were proposed.
Complementary benchmarks explore additional SWE tasks, including the ability to generate and validate tests from GitHub issues~\cite{ahmed2024tdd, mundler2024swt}, or solving real-world IT automation tasks~\cite{jha2025itbench}. Notably, Terminal-Bench~\cite{merrill2025terminal} focuses on interactive terminal usage, assessing agents’ command-line proficiency.

More recent benchmarks push toward more challenging and realistic evaluation. SWE-Lancer~\cite{miserendino2025swe} collected 1,400 freelance tasks from Upwork with total payouts over $\$1M$. Tasks include both technical fixes and managerial decisions, evaluated via verified tests or comparison to human managers' choices. Results expose gaps in long-term reasoning and decision-making. Similarly, SWE-bench Pro~\cite{deng2025swebenchproaiagents} introduces 1,865 human-verified tasks spanning 41 repositories, often requiring multi-file edits and hours of effort. Model performance remains below $25\%$ Pass@1, highlighting current limitations in handling long-horizon, complex code changes.





\subsection{Scientific Agents} \label{sec:sa}

Scientific agents automate core research tasks by integrating domain knowledge and scientific tools.
Their evaluation has progressed from basic reasoning benchmarks to comprehensive frameworks assessing diverse scientific research capabilities. Early efforts focused on knowledge recall and reasoning~\cite{clark2018think, lu2022learnexplainmultimodalreasoning, wang-etal-2022-scienceworld}, and literature understanding~\cite{dasigi2021datasetinformationseekingquestionsanswers,qasa, deyoung2021ms2multidocumentsummarizationmedical}. More recent benchmarks like SciRiff~\cite{wadden2024sciriff} broaden the scope to instruction-following across scientific domains.


Recent advancements have shifted the focus toward developing and assessing scientific agents in accelerating scientific research. New benchmarks span the full research pipeline:
(1) Scientific Ideation: Evaluates agents’ ability to generate novel, expert-level research ideas, emphasizing creativity, relevance, and scientific feasibility~\cite{si2025can}.
(2) Experiment Design: Benchmarks like AAAR-1.0~\cite{lou2025aaar} assess systematic experiment planning, hypothesis formulation, appropriate use of methodologies, and rigor of experimental procedures.
(3) Code Generation: Benchmarks such as SciCode~\cite{tian2024scicode}, ScienceAgentBench~\cite{chen2025scienceagentbench}, CORE-Bench~\cite{siegel2024corebenchfosteringcredibilitypublished}, and PaperBench~\cite{starace2025paperbenchevaluatingaisability} test the agent's ability to produce accurate, executable scientific code~\cite{chan2025mlebenchevaluatingmachinelearning}.
(4) Peer Review: Evaluates whether agents can generate substantive, high-quality reviews~\cite{chamoun-etal-2024-automated}.
Ultimately, the field is moving toward benchmarks that encompass the full research cycle, leading to the evaluation of innovative scientific discovery.

\subsection{Conversational Agents} \label{sec:conv}

Conversational agents are agents designed to perform goal-directed, multi-turn dialogue with a user to accomplish a specific task, such as booking a flight or resolving a customer support issue.
This field builds on Task-Oriented Dialogue Systems (TODS), extending it from purely textual tasks to real-world tasks requiring environment interaction.



Early TODS benchmarks focused on multi-turn conversations across different domains~\cite{budzianowski-etal-2018-multiwoz, andreas-etal-2020-task}, and distinct user intents~\cite{chen-etal-2021-action}.
Yet these benchmarks were confined to textual interaction. To address that, Sierra proposed $\tau$-Bench~\cite{yao2024taubenchbenchmarktoolagentuserinteraction}. It assesses agents' ability to interact with simulated users to accomplish their tasks by effectively utilizing API tools, while adhering to domain-specific policies. $\tau$-Bench spans domains like retail and airline customer service. However, it is limited in scale, user simulation setup, and focuses solely on coarse-grained end-to-end metrics, overlooking policy violations and dialogue flow errors.
$\tau^2$-Bench~\cite{barres2025tau2benchevaluatingconversationalagents} addresses these limitations by introducing a Telecom domain, where the user utilizes tools to act in a shared, dynamic environment, and by adding a compositional task generator that programmatically creates diverse, verifiable tasks.

Although $\tau$-Bench is the most dominant benchmark in this category, it builds on prior work that automates benchmark creation and enables dynamic evaluation.
IntellAgent~\cite{levi2025intellagentmultiagentframeworkevaluating} proposes an automated process for generating synthetic test scenarios from database schemas and policy documents. 
They showed a high correlation between $\tau$-Bench and their synthetic benchmark results.
ALMITA~\citep{arcadinho-etal-2024-automated} used a hybrid approach that starts from user intents and a sequence of intermediate LLM-generated graphs, followed by manual filtering to generate diverse, realistic customer support scenarios.




\section{Generalist Agent Evaluation} \label{sec:general}
Similar to the evolution of LLMs from task-specific into general-purpose models, agents are also transitioning from application-specific toward general-purpose agents~\cite{bandel2026agentic}. Such tasks require integrated capabilities, from planning, reasoning, and tool use to web interaction, file handling, code execution, and more.
This shift has led to the development of two complementary approaches to generalist agent evaluation.

One approach aims to address this gap by proposing benchmarks that inherently require a wide range of capabilities.
A key benchmark in this category is Gaia~\cite{mialon2023gaia}. Gaia is composed of real-world questions that require abilities such as reasoning, multi-modality handling, web browsing, and tool-use proficiency.
Over time, model performance on Gaia improved, saturating the easier portion of the benchmark and emphasizing the need for an updated version.
\citet{andrews2025arescalingagentenvironments} proposed Gaia2 as a mobile environment with apps such as
email, messaging, and calendar. The set of required capabilities was extended to include handling ambiguity, noise, temporal constraints, and multi-agent collaboration. 

This work continues a class of benchmarks that evaluate agents in full computer environments. OSWorld~\cite{xie2024osworld}, AppWorld~\cite{trivedi-etal-2024-appworld}, and more~\cite{omniact, bonatti2025windows} test the agent's ability to execute complex tasks across applications, manage control flows, and ensure stable performance in real settings. They differ in the way the agent interacts with the environment. In OSWorld, the agent interacts via UI, while in AppWorld and Gaia2, it uses code and API calls.

The second approach for generalist agent evaluation relies on unifying several task-specific benchmarks into one.
AgentBench~\cite{liu2023agentbench} introduces interactive environments spanning OS operations, databases, games, and household tasks—highlighting core skills like flexibility and tool-based problem solving.
HAL (Holistic Agent Leaderboard;~\citealp{hal}) provides a unified platform for benchmarks across domains, including coding and web. 
Yet, this approach does not support the same agent harness evaluation across any benchmark environment. Harbor and Exgentic aim to solve that by providing frameworks with a unified protocol for general agent assessment~\cite{Harbor_Framework, bandel2026generalagentevaluation}.
These represent the first steps toward holistic, standardized, cross-environment agent evaluation~\cite{lacoste2026cubestandardunifyingagent}.

\section{Core Benchmark Dimensions}\label{sec:bench_dims}


The preceding sections reviewed benchmarks organized by agent type and application domain. Here, we take a broader perspective and analyze benchmarks along shared orthogonal dimensions: data curation, environment, interface, metric, and safety. This analysis reveals common structural patterns and gaps.
In Table \ref{tab:benchmarks_analysis}, we compare representative benchmarks based on these dimensions.


\paragraph{Data Curation}
High-quality benchmarks rarely rely solely on human-curated data; most employ a \textit{hybrid curation} strategy. For example, SWE-bench Verified refines the original harvest of GitHub issues through human validation to improve the robustness and reliability of the benchmark.
Similarly, AppWorld constructs tasks using a synthetic "world" of 100 fictitious users but validates them with programmatic checks. Other benchmarks, like Mind2Web, derive data from real-world interaction logs, which are then cleaned and annotated. In contrast, GAIA specifically utilizes humans for crafting and validating the questions, ensuring challenging, unambiguous questions that are conceptually simple for humans but require complex tool usage for agents.
This raises the tension between the reliance on human annotation for ensuring data validity and the need to automate the data curation and validation process, allowing scalable and adaptable evaluation. This tension emphasizes the need for methods that automate this process without compromising its quality.

\paragraph{Environment}
Evaluation environments generally fall into two categories: \textit{static} and \textit{dynamic}. Static environments, such as the original Mind2Web, rely on offline traces or cached web pages where agents predict the next action without influencing the state. 
While scalable, these fail to capture the cascading effects of errors: 
incorrect actions have no downstream consequences, missing compounding failures that can ultimately cause task failure.
Dynamic environments, conversely, allow agents to interact with a live or simulated world (e.g., Docker containers for SWE, browser sandboxes for web agents), where their actions alter the state the agent observes, enabling diagnosis of failure modes in long-horizon tasks.
As agents become more capable, benchmarks must embrace dynamic environments that assess the agent's ability to evaluate long-horizon interactions with the environment.


\paragraph{Interaction Interface}
The interface defines the communication protocol between the agent and the environment, governing the action and observation spaces. We categorize benchmarks into three primary interfaces:
\textit{Code and Terminal} interface requires agents to generate executable scripts, e.g., Python, Bash, or SQL. This interface is predominant in SWE and scientific benchmarks, where success depends on logic and syntactic correctness.
\textit{Tools} interface restricts actions to predefined function calls. Benchmarks such as $\tau$-Bench and AppWorld rely on tool calling with schema adherence to achieve their goal.
\textit{Graphical User Interfaces (GUI)} simulate human interaction via accessibility trees (e.g., HTML DOM) or visual UI. Web and consumer-facing applications benchmarks utilize this modality to evaluate visual grounding and computer or web navigation.

\paragraph{Metric} The most ubiquitous metric is task completion, yet its implementation varies depending on the application and expected output. For SWE tasks, where the output is a piece of functional code, SWE measures are adopted, like execution-based \textit{unit testing}. 
Tasks that require modifying the environment state, like $\tau$-Bench, deploy \textit{state matching} against the gold state to assess the modification correctness. 
For multi-step reasoning tasks, such as GAIA, \textit{answer matching} is used to verify unambiguous short-form responses against a gold standard.
This metric divergence emphasizes the need for targeted verification steps to ensure the benchmark's validity \citep{zhu2025establishing}. 
Notably, such binary outcome metrics are insufficient to understand the intermediate agent's progress, and call for fine-grained evaluation solutions (See \S\ref{sec:dis}).

\paragraph{Safety and Robustness Metrics}
While most benchmarks prioritize capability, safety, and robustness are critical for enterprise adoption. Robustness is often quantified via pass\textasciicircum k, the fraction of tasks where the agent succeeds across all k independent runs. Beyond robustness, enterprise agents must adhere to strict policies, such as data privacy and access control, which are rarely tested in standard benchmarks. For instance, SWE-Lancer does not inherently penalize risky behaviors unless they interfere with replicating the target behavior. Future benchmarks must integrate "guardrail" metrics, penalizing agents that achieve task success via non-compliant actions (e.g., deleting production databases).

\begin{table}[t!]
\centering
\resizebox{\columnwidth}{!}{%
\begin{tabular}{@{}l c c c c c@{}}
\toprule
\textbf{Benchmark} & \textbf{Data} & \textbf{Env.} & \textbf{Interface} & \textbf{Metric} & \textbf{Safety} \\
\midrule
\textbf{SWE-bench Ver.} & Hybrid & Dynamic & Code & Unit Tests & No \\

\textbf{SWE-Lancer} & Hybrid & Dynamic & Code & End-to-end & No \\

\textbf{Mind2Web} & Hybrid & Static & GUI & Action Match & No \\

\textbf{WebArena} & Hybrid & Dynamic & GUI & Mix & No \\

\textbf{PaperBench} & Hybrid & Dynamic & Code & End-to-end & No \\

\textbf{TAU-Bench} & Hybrid & Dynamic & Tools & State Match & Yes \\

\textbf{AppWorld} & Hybrid & Dynamic & Tools & State Match & No \\

\textbf{GAIA} & Human & Dynamic & Mix & Answer Match & No \\
\bottomrule
\end{tabular}%
}
\caption{Comparative analysis of representative agent benchmarks. We characterize each benchmark by its data curation strategy, environment dynamicity, interaction interface, evaluation metric, and whether it explicitly incorporates safety constraints.}
\label{tab:benchmarks_analysis} 
\end{table}

\section{Frameworks for Agent Evaluation}
\label{sec:frameworks}

To meet the growing need for systematic assessment of LLM agents, several general-purpose frameworks have emerged, offering developers tools for continuous monitoring, evaluation, error analysis, and performance optimization. Unlike the benchmarks discussed in the preceding sections, which assess fully developed systems using fixed scenarios and standardized test datasets, these frameworks integrate into the development process, enabling flexible, custom scenario design and supporting a broad range of general use cases across both development and deployment.



\input{tables/framworks}

There are many frameworks supporting the evaluation of a wide range of agent types,
including LangSmith~\citep{langsmith}, Langfuse~\citep{langfuse}, Google Vertex AI evaluation service ~\citep{vertex_ai_gen_ai_evaluation}, Arize AI's Evaluation Framework~\citep{arize_ai_evaluation_framework}, Galileo Agentic Evaluation~\citep{galileo_agentic_evaluations}, Patronus AI \citep{patronus_ai}, W\&B Weave~\citep{wandbevaluations2023}, LangChains' AgentEvals \citep{agentevals}; Databricks Mosaic AI Agent Evaluation~\citep{mosaic_ai_agent_evaluation}, which is mostly designed for RAG like tasks; 
Botpress Multi-Agent Evaluation System \citep{Kargwal2025} and AutoGen \citep{dibia2024autogen} for multi-agent systems; and more. 

All evaluation platforms provide continuous monitoring of agent trajectories, assessing key performance metrics such as task completion rates, latency, execution speed, and, in some cases, throughput and memory usage~\cite{langsmith}. Some frameworks utilize the OpenTelemetry~\citep{blanco2023practical} observability framework and their infrastructure, including Langfuse and Google Vertex AI. 

Beyond observability, each framework applies quality assessment methods across multiple levels of granularity: 

\textbf{Final Response Evaluation.}
Frameworks often incorporate LLM-based judges to evaluate agent responses against predefined criteria (such as faithfulness or politeness), with some offering proprietary judge models (e.g., Databricks Mosaic and PatronusAI). Additionally, most platforms allow for customizable assessment metrics, enabling domain-specific evaluation of output quality and relevance. Final-response evaluation is fast, inexpensive, and easy to automate, making it well-suited for large-scale monitoring and regression testing. However, it offers limited insight into agent behavior, as it cannot assess intermediate decisions, execution efficiency, or failure causes within complex workflows. 

\textbf{Stepwise Evaluation.}
Most frameworks support granular assessments of individual agent steps, such as LLM generations, tool invocations, and routing decisions, enabling error localization and systematic analysis of where and how multi‑step executions fail. 
A common approach evaluates each step independently using predefined or customizable judges, often LLM‑based or lightweight classifiers, that assess quality attributes such as correctness, relevance, or instruction adherence. 
Additionally, many frameworks perform tool‑specific evaluation by validating tool choice, parameter schemas, and output usability. To better align evaluation with agent structure, Arize Phoenix provides agent‑specific step templates, which tailor evaluation criteria to particular stages such as routing, planning, retrieval, or reflection. 

These approaches assume that each action can be meaningfully assessed in isolation, overlooking dependencies between steps.
To address this gap, Galileo Agentic Evaluation incorporates a goal‑progress‑oriented \emph{action advancement metric}, which measures whether each step successfully contributes to or advances toward a user-defined goal. 

\textbf{Trajectory-Based Assessment.}
Some platforms, such as Google Vertex AI and LangSmith, also support trajectory-based assessments, which move beyond individual steps and analyze how an agent navigates toward task completion. Current approaches fall into two broad categories.
\textbf{\textit{Reference-Based}} methods compare the trajectory against an expected optimal path, measuring alignment between the observed and gold action sequences. Platforms such as LangSmith, Vertex AI, and AgentEvals support various alignment modes, including exact, partial, unordered, and subset matching. 
AgentEvals further extends this through graph-based evaluation: for frameworks that model agents as graphs, it assesses whether the agent visits the expected nodes and transitions rather than requiring alignment over a flat sequence of tool calls. However, reference-based methods are inherently limited, as multiple valid paths typically exist and manual reference specification is often infeasible.
\textbf{\textit{Reference-Free}} methods use LLM-based judges to evaluate trajectory quality without a predefined gold path, assessing properties such as coherence, efficiency, or goal directedness directly from the observed sequence.

A core methodological consideration is the choice between reference-based and reference-free evaluation. Reference-based methods offer precision and reproducibility but depend on predefined expected behavior. Reference-free methods, typically relying on LLM-based judges, provide greater flexibility at the cost of reliability. This tension extends to judge design: general-purpose judges offer broad coverage but lower precision, while task-specific judges excel on their target criteria at the cost of a narrower scope.


\textbf{Supporting Capabilities}. Beyond quality assessment, frameworks provide supporting capabilities for data management and experimentation. Most offer integrated annotation tools and support \textit{human-in-the-loop} evaluation, enabling the extraction of evaluation datasets from production logs. Platforms such as Patronus AI and Databricks Mosaic additionally facilitate \textit{synthetic data generation} using proprietary seed data. Current frameworks also support \textit{A/B comparisons}, enabling side-by-side analysis of inputs, outputs, and metrics across runs, and in some cases (e.g., Patronus AI) of aggregated results across experimental setups.



Table~\ref{tab:agent-evaluation-frameworks} presents key frameworks for agent evaluation along with their support for the evaluation features discussed in this section.

Overall, current evaluation frameworks show great value but still face several open challenges.
First, while analyzing individual traces is well supported, deriving insights across large collections of runs remains difficult. Thus, understanding root causes for agent failures at scales is still unsupported. A/B comparisons partially address this by allowing side-by-side analysis of different experimental setups. Yet, causally attributing outcome differences to specific steps or decisions is not yet possible. 
Second, current frameworks overlook the cost incurred by the evaluation process itself, specifically when scaling over a large number of traces with expensive LLM judges. This emphasizes the need for more efficient evaluation processes, with better resource allocation techniques.
Finally, most current frameworks lack built-in support for evaluating safety and policy compliance (See \S\ref{sec:dis}).
\\
\\
\textbf{Gym-like Environments.} 

Gym-like Environments provide controlled, interactive environment simulations for agent evaluation. Inspired by OpenAI Gym \citep{brockman2016openai}, originally designed for training and evaluating Reinforcement Learning algorithms, these frameworks have been adapted to the training and evaluation of LLM agents using realistic task simulations, allowing them to interact with dynamic environments. 
Moreover, these frameworks 
enable standardized evaluation across various benchmarks, with environments made for web agents \citep{chezelles2024browsergym}, AI research agents \citep{nathani2025mlgym}, and SWE agents \citep{pan2024training}.

\section{Discussion} \label{sec:dis}

\subsection{Current Trends} \label{sec:cur}
Our review identifies two key trends shaping the current landscape of agent evaluation:

\textbf{Realistic and Challenging Evaluation.}
While early evaluations often employed simplified, static environments, there is a clear shift toward more realistic and complex benchmarks. Web agents evolved from basic simulations to dynamic, real-world settings such as \textit{WebArena}. In software engineering, benchmarks like \textit{SWE-bench} utilize real GitHub issues, moving beyond synthetic coding tasks. Furthermore, there is increasing interest in long-horizon tasks typically performed by highly trained human experts, pushing evaluation closer to real-world professional workflows.
This change reflects the growing need to assess agents' ability to derive practical value in realistic settings, to stress-test their limitations, and guide their progress.

\textbf{Live Benchmarks.}
The rapid pace of LLM and agent development necessitates adaptive and continuously updated evaluation methods. Static benchmarks quickly become obsolete, saturated, and abandoned.
In response, we see a rise in ``live'' benchmarks. For instance, the few versions of \textit{BFCL}, and the refinement of the \textit{SWE-bench} family (\textit{Verified}, \textit{Pro}, and more). 
These ongoing adaptations serve several purposes: matching the increased capabilities of agents, addressing shortcomings of previous versions (e.g., by conducting human verification and refining success metrics), and adapting to evolving research ecosystems such as MCP-based tool calling.
This dynamic benchmarking approach is essential to maintain relevance in a fast-moving field.

\subsection{Future Directions} \label{sec:emergent}
We also recognize critical gaps in current agent evaluation that future research must address.

\textbf{Advancing Granular Evaluation.}
Many current benchmarks rely on coarse-grained, end-to-end success metrics that, while useful for gauging overall performance, fall short in diagnosing specific agent failures. This lack of granularity obscures insights into intermediate decision processes such as tool selection and reasoning quality. Addressing this limitation calls for the development of standardized, fine-grained evaluation metrics targeting the trajectory of an agent’s task execution. Future work should explore detailed, step-by-step assessments
to provide richer feedback and guide targeted improvements.

\textbf{Cost and Efficiency Metrics.}
Current evaluations often prioritize performance while overlooking cost and efficiency measurements. This emphasis can inadvertently drive the development of highly capable but resource-intensive agents, limiting their practical deployment~\citet{kapoor2024ai}. Future evaluation frameworks should integrate cost efficiency as a core metric, tracking factors such as token usage, API expenses, inference time, and overall resource consumption. Establishing standardized cost metrics will help guide the development of agents that balance performance with operational viability.

\textbf{Scaling \& Automating.}
The reliance on static, human-annotated evaluation data poses significant scalability challenges. It is resource-intensive, and the resulting benchmarks quickly become outdated in this rapidly evolving field. This shortcoming underscores the need for scalable, automated evaluation, which may be addressed via \textit{synthetic data generation} techniques. 
Another avenue is automating evaluation by employing an LLM or agent as a Judge~\citet{zhuge2024agent}.
This approach not only reduces the reliance on resource-intensive human annotation but also holds the potential to capture more nuanced aspects of agent performance.

\textbf{Safety and Compliance.}
Current benchmarks lack sufficient focus on safety,
trustworthiness, and policy compliance. While early efforts
have begun to address these dimensions, evaluations still lack comprehensive tests for robustness against adversarial inputs, bias mitigation, and organizational and societal policy compliance.
Future research should include safety metrics in benchmarks as well as developing safety benchmarks that simulate real-world scenarios, particularly in multi-agent scenarios where emergent risks may arise \cite{hammond2025multi}. This can ensure that agents are not only effective but also safe and secure.

\textbf{Decoupling LLM \& Harness Evaluation.}
Most current agent benchmarks conflate two distinct evaluation targets: (1) the inherent capabilities of the backbone LLM, and (2) the design of the agent Harness (a.k.a. scaffold). Disentangling these is essential for enabling systemic attribution of performance gains. Efforts like Harbor and Exgentic begin to address this by standardizing agent evaluation across models and agent harness settings. 
Future work should develop controlled evaluation protocols that vary each factor independently, enabling systematic comparison and isolating the contribution of individual components, whether LLM capabilities, harness design, or specific modules such as memory or planning, to overall agent performance.



\section{Conclusion}
This survey presents an overview of the evolving field of LLM-based agent evaluation, outlining its progression from assessing isolated capabilities in simplified settings to evaluating agents in realistic, dynamic, and 
challenging environments. While notable progress has been made, several challenges remain.
Future research should focus on developing fine-grained, scalable evaluation methods that extend beyond overall success rates, establishing standardized metrics for cost-efficiency, safety, and robustness for responsible deployment.
For practical guidance, we provide actionable benchmark recommendations in Appendix \S\ref{app:recommendations}.


\section*{Limitations}
\label{sec:limitations}
While this survey provides a comprehensive overview of the evaluation landscape for LLM-based agents, it is important to acknowledge certain limitations inherent to a review of such a rapidly evolving field.

First, the domain of LLM-based agents and their evaluation is exceptionally dynamic. New benchmarks, evaluation frameworks, agent architectures, and research findings are being published at an unprecedented pace. Although we have striven to incorporate the most current and impactful work up to the time of writing, this survey inevitably represents a snapshot. Some very recent or forthcoming developments might not have been included. Our commitment to maintaining a continuously updated GitHub repository, as mentioned in the abstract, aims to mitigate this challenge over time for the community.

Second, the selection of benchmarks and frameworks, while intended to be representative, is subject to the breadth of the field. To maintain clarity and focus, we prioritized works that illustrate key trends or address significant aspects of agent evaluation. Consequently, some specific or niche evaluation approaches may not have received detailed coverage. 

Third, in covering a wide array of topics, the depth of analysis for each individual benchmark or framework is necessarily constrained. Readers requiring an exhaustive understanding of a particular evaluation tool or methodology are encouraged to consult the primary research articles cited throughout this survey.


Finally, the discussion on ``Future  Directions'' (\S\ref{sec:emergent}) and the identification of ``critical gaps'' inherently involve a degree of foresight and interpretation based on current trends. While these are informed by our analysis of the existing literature, the actual trajectory of future research and the relative importance of these identified areas will continue to evolve.

Despite these limitations, we believe this survey offers a valuable and structured synthesis of the current state of LLM-based agent evaluation, serving as a useful resource for researchers, developers, and practitioners in the field.


\bibliography{custom}

\appendix
\input{appendix}

\end{document}

%% file: commands.tex
\usepackage[export]{adjustbox}

\usepackage{subfig}

\usepackage{booktabs}
\usepackage{hyperref}
\usepackage{longtable}
\usepackage{tcolorbox}

\usepackage{bbm}
\usepackage{stmaryrd}
\usepackage{amsmath,amssymb}

\usepackage{amsthm}

\usepackage{refcount}


%% file: figures/layout_fig_short.tex
\definecolor{paired-light-blue}{RGB}{198, 219, 239}
\definecolor{paired-dark-blue}{RGB}{49, 130, 188}
\definecolor{paired-light-orange}{RGB}{251, 208, 162}
\definecolor{paired-dark-orange}{RGB}{230, 85, 12}
\definecolor{paired-light-green}{RGB}{199, 233, 193}
\definecolor{paired-dark-green}{RGB}{49, 163, 83}
\definecolor{paired-light-purple}{RGB}{218, 218, 235}
\definecolor{paired-dark-purple}{RGB}{117, 107, 176}
\definecolor{paired-light-gray}{RGB}{217, 217, 217}
\definecolor{paired-dark-gray}{RGB}{99, 99, 99}
\definecolor{paired-light-pink}{RGB}{222, 158, 214}
\definecolor{paired-dark-pink}{RGB}{123, 65, 115}
\definecolor{paired-light-red}{RGB}{231, 150, 156}
\definecolor{paired-dark-red}{RGB}{131, 60, 56}
\definecolor{paired-light-yellow}{RGB}{231, 204, 149}
\definecolor{paired-dark-yellow}{RGB}{141, 109, 49}
\definecolor{light-green}{RGB}{118, 207, 180}
\definecolor{raspberry}{RGB}{228, 24, 99}

\tikzset{%
    root/.style =          {align=center,text width=3cm,rounded corners=3pt, line width=0.5mm, fill=paired-light-gray!50,draw=paired-dark-gray!90},
    data_section/.style =  {align=center,text width=4cm,rounded corners=3pt, fill=paired-light-blue!50,draw=paired-dark-blue!80,line width=0.4mm},
    model_section/.style = {align=center,text width=4cm,rounded corners=3pt, fill=paired-light-orange!50,draw=paired-dark-orange!80,line width=0.4mm},
    training_section/.style = {align=center,text width=4cm,rounded corners=3pt, fill=paired-light-green!50,draw=paired-dark-green!80, line width=0.4mm},
    inference_section/.style = {align=center,text width=4cm,rounded corners=3pt, fill=paired-light-red!35,draw=paired-light-red!90, line width=0.4mm},
    discussion_section/.style = {align=center,text width=4cm,rounded corners=3pt, fill=paired-light-purple!35,draw=paired-dark-purple!90, line width=0.4mm},
    subsection/.style =    {align=center,text width=3.5cm,rounded corners=3pt}, 
}

\begin{figure*}[!htb]
    \centering
    \resizebox{1\textwidth}{!}{
    \begin{forest}
        for tree={
            forked edges,
            grow'=0,
            draw,
            rounded corners,
            node options={align=center},
            text width=4cm,
            s sep=6pt,
            calign=child edge,
            calign child=(n_children()+1)/2,
            l sep=12pt,
        },
        [Agent Evaluation, root, , calign child=2,
            [Agent Capabilities Evaluation (\S\ref{sec:capabilities}), data_section
                [Planning and Multi-Step Reasoning (\S\ref{sec:plan}.1), data_section 
                    [ 
                    \textit{PlanBench} ~\cite{valmeekam2023planbench};
                    FlowBench~\cite{xiao2024flowbench};
                    \textit{Natural Plan} ~\cite{zheng2024natural}
                    ,data_section, text width=12cm
                    ] 
                ]
                [Function Calling \& Tool Use (\S\ref{sec:tool}.2), data_section
                    [\textit{BFCL} \cite{berkeley-function-calling-leaderboard};
                    \textit{ToolBench} \cite{qin2023toolllm};
                    \textit{ComplexFuncBench} \cite{zhong2025complexfuncbench};
                    ,
                    data_section, text width=12cm
                    ] 
                ]
                [Self-Reflection (\S\ref{sec:self_reflect}.3), data_section
                    [
                    \textit{LLM-Evolve}~\citep{you-etal-2024-llm};
                    \textit{LLF-Bench} \citep{cheng2023llfbenchbenchmarkinteractivelearning};
                    ,
                    data_section, text width=12cm
                    ] 
                ]
                [Memory (\S\ref{sec:memory}.4), data_section
                    [
                    \textit{StreamBench} ~\cite{wu2024streambench};
                    MemBench~\cite{tan-etal-2025-membench};
                    MemoryAgentBench~\cite{hu2025evaluating};
                    ,
                    data_section, text width=12cm
                    ]
                ]
            ]
            [Application-Specific Agent Evaluation (\S\ref{sec:use_case}), model_section
                [Web Agents (\S\ref{sec:web}), model_section 
                    [ 
                    \textit{WebShop} \cite{yao2022webshop};
                    \textit{Mind2web} \cite{deng2023mind2web};
                    \textit{WebArena} \citep{zhou2023webarena};
                    \textit{AssistantBench} \cite{yoran2024assistantbench};
                    \textit{ST-WebAgentBench} \cite{levy2024st};
                    ,
                    model_section, text width=12cm] 
                ]
                [Software Engineering \\ Agents (\S\ref{sec:se}), model_section
                    [
                    \textit{SWE-bench}~\cite{Jimenez2023SWEbenchCL};
                    \textit{SWE-bench Verified}~\cite{swebenchverified2024};
                    \textit{SWELancer} \cite{miserendino2025swe},
                    model_section, text width=12cm] 
                ]
                [Scientific Agents (\S\ref{sec:sa}), model_section
                    [
                    \textit{ScienceAgentBench} \citep{chen2025scienceagentbench};
                    \textit{CORE-Bench} \citep{siegel2024corebenchfosteringcredibilitypublished};
                    PaperBench~\cite{starace2025paperbenchevaluatingaisability};
                    \textit{SciCode} \citep{tian2024scicode};
                    ,
                    model_section, text width=12cm] 
                ]
                [Conversational Agents (\S\ref{sec:conv}), model_section
                    [
                    \textit{$\tau$-Bench} \cite{yao2024taubenchbenchmarktoolagentuserinteraction};
                    $\tau^2$-Bench~\cite{barres2025tau2benchevaluatingconversationalagents};
                    \textit{IntellAgent} \cite{levi2025intellagentmultiagentframeworkevaluating};
                    \textit{ALMITA} \cite{arcadinho-etal-2024-automated};
                    , model_section, text width=12cm] 
                ]
            ]
            [Generalist Agents Evaluation (\S\ref{sec:general}), training_section
                [, training_section
                [
                \textit{Gaia2}~\cite{andrews2025arescalingagentenvironments};                
                \textit{OSWorld}~\cite{xie2024osworld};
                \textit{AppWorld}~\cite{trivedi-etal-2024-appworld};
                \textit{TheAgentCompany}~\cite{xu2024theagentcompany};
                \textit{AgentBench}~\cite{liu2023agentbench};
                \textit{HAL}~\cite{hal},
                training_section, text width=12cm]
                ]
            ]
            [Frameworks for Agent Evaluation
            (\S\ref{sec:frameworks}), inference_section
                [Development Frameworks, inference_section
                [
                \textit{LangSmith} \citep{langsmith};
                \textit{Langfuse} \citep{langfuse};
                 \textit{Galileo} \citep{galileo_agentic_evaluations};
                \textit{Vertex AI} \citep{vertex_ai_gen_ai_evaluation};
                ,
                inference_section, text width=12cm]
                ]
                [Gym-like Environments, inference_section
                [
                \textit{MLGym}\citep{nathani2025mlgym};
                \textit{BrowserGym}\citep{chezelles2024browsergym};
                \textit{SWE-Gym}\citep{pan2024training},
                inference_section, text width=12cm]
                ] 
            ]
        ]
    \end{forest}
    }
    \caption{Overview of the paper with core works.
    } 
    \label{fig:agent-evaluation-typology} 
\end{figure*}

%% file: tables/framworks.tex
\begin{table*}[htbp] 
\centering 
\resizebox{\textwidth}{!}{
\begin{tabular}{lcccccc} 
\toprule
Framework & Stepwise Assessment & Monitoring & Trajectory Assessment & Human in the Loop & Synthetic Data Generation & A/B Comparisons \\ 
\midrule
LangSmith (\citeauthor{langsmith}) & 	\textcolor{teal}{\checkmark} & 	\textcolor{teal}{\checkmark} & 	\textcolor{teal}{\checkmark} & 	\textcolor{teal}{\checkmark} & \textcolor{purple}{\bm$\times$} & 	\textcolor{teal}{\checkmark} \\ 
Langfuse (\citeauthor{langfuse}) & 	\textcolor{teal}{\checkmark} & 	\textcolor{teal}{\checkmark} & \textcolor{purple}{\bm$\times$} & 	\textcolor{teal}{\checkmark} & \textcolor{purple}{\bm$\times$} & 	\textcolor{teal}{\checkmark} \\ 
Google Vertex AI
evaluation (\citeauthor{vertex_ai_gen_ai_evaluation}) & 	\textcolor{teal}{\checkmark} & 	\textcolor{teal}{\checkmark} & 	\textcolor{teal}{\checkmark} & \textcolor{purple}{\bm$\times$} & \textcolor{purple}{\bm$\times$} & 	\textcolor{teal}{\checkmark} \\ 
 Arize AI’s Evaluation (\citeauthor{arize_ai_evaluation_framework}) & 	\textcolor{teal}{\checkmark} & 	\textcolor{teal}{\checkmark} & \textcolor{purple}{\bm$\times$} & 	\textcolor{teal}{\checkmark} & 	\textcolor{teal}{\checkmark} & 	\textcolor{teal}{\checkmark} \\ 
Galileo Agentic Evaluation (\citeauthor{galileo_agentic_evaluations}) & 	\textcolor{teal}{\checkmark} & 	\textcolor{teal}{\checkmark} & \textcolor{purple}{\bm$\times$} & 	\textcolor{teal}{\checkmark} & \textcolor{purple}{\bm$\times$} & 	\textcolor{teal}{\checkmark} \\ 
Patronus AI (\citeauthor{patronus_ai}) & 	\textcolor{teal}{\checkmark} & 	\textcolor{teal}{\checkmark} & \textcolor{purple}{\bm$\times$} & 	\textcolor{teal}{\checkmark} & 	\textcolor{teal}{\checkmark} & 	\textcolor{teal}{\checkmark} \\ 
AgentsEval (\citeauthor{agentevals}) & \textcolor{purple}{\bm$\times$} & \textcolor{purple}{\bm$\times$} & 	\textcolor{teal}{\checkmark} & \textcolor{purple}{\bm$\times$} & \textcolor{purple}{\bm$\times$} & \textcolor{purple}{\bm$\times$} \\ 
Mosaic AI (\citeauthor{mosaic_ai_agent_evaluation}) & 	\textcolor{teal}{\checkmark} & 	\textcolor{teal}{\checkmark} & \textcolor{purple}{\bm$\times$} & 	\textcolor{teal}{\checkmark} & 	\textcolor{teal}{\checkmark} & 	\textcolor{teal}{\checkmark} \\ 
\bottomrule
\end{tabular}}
\caption{Supported evaluation capabilities of major agent frameworks. Note that some of these capabilities are still in initial phases of development, as discussed further in the text.}
\label{tab:agent-evaluation-frameworks} 
\end{table*}

%% file: appendix.tex
\newpage
\section{Literature Review Methodology}
\label{app:literature_review}

To ensure the comprehensiveness and validity of this survey, we employed a rigorous, multi-stage literature review process designed to capture the rapidly evolving landscape of LLM agent evaluation. Our methodology consisted of three primary phases: systematic search, structured selection, and expert validation.

\paragraph{Search Strategy and Citation Chasing}
Our initial data collection involved a systematic search across major academic repositories and archives, including Google Scholar, the ACL Anthology, HuggingFace Papers, and arXiv. We utilized a targeted set of keywords and their combinations, such as ``LLM agent evaluation,'' ``agent benchmark,'' ``tool use evaluation,'' and ``web agent benchmark.''
Following the identification of seminal and highly cited works, we applied an iterative ``snowballing'' technique. We performed both forward and backward citation chasing on these seed papers to uncover foundational prior art as well as emerging methodologies that might not yet be indexed by standard keywords.

\paragraph{Inclusion and Exclusion Criteria}
We defined strict criteria to maintain the survey's focus and quality. Papers were \textit{included} if they introduced a novel benchmark, an evaluation framework, or a significant methodological contribution to the assessment of LLM-based agents.
Conversely, we \textit{excluded} works that: 1. Proposed new agent architectures without a distinct contribution to evaluation methodology. 2. Focused solely on traditional LLM evaluation (e.g., reasoning on static datasets), lacking dynamic, agentic, or interactive components.
While acceptance to top-tier conferences was utilized as a primary indicator of quality, we also included high-impact preprints to ensure the survey reflects the most recent advancements in this fast-paced field.

\paragraph{Expert Consultation}
To mitigate coverage gaps and ensure an accurate representation of specific subdomains, we consulted with domain experts corresponding to the key categories in our taxonomy (e.g., SWE and web agent researchers). We specifically engaged with creators of leading benchmarks and dominant agent architectures to discuss the selected manuscripts. This validation step ensured that our survey captures the nuances of state-of-the-art evaluation protocols.

\section{Agent Capabilities Evaluation}
\label{app:capabilities}
In this appendix, we expand on the evaluation of LLM-based agent capabilities. While the main paper presents the core information, here we offer a more detailed account of each benchmark, its interrelations, and how the evaluation of each capability evolves.

\input{figures/layout_figs}

\subsection{Planning and Multi-Step Reasoning}\label{app:plan}

Planning and multi-step reasoning form the foundation of an LLM agent's ability to tackle complex tasks effectively which enables agents to decompose problems into smaller, more manageable subtasks and create strategic execution paths toward solutions \cite{gao2023largelanguagemodelsempowered}.

Multi-step reasoning in LLMs typically involves executing sequential logical operations—typically requiring 3-10 intermediate steps—to arrive at solutions that cannot be derived through single-step inference ~\cite{cobbe2021gsm8k,yang2018hotpotqadatasetdiverseexplainable,suzgun2022bbh}. This foundational need for multi-step planning has led to the development of specialized benchmarks and evaluation frameworks that systematically assess these capabilities across diverse domains, including: mathematical reasoning (GSM8K ~\cite{cobbe2021gsm8k}, MATH ~\cite{hendrycks2021MATH}, AQUA-RAT ~\cite{ling2017AQUA}), multi-hop question answering (HotpotQA ~\cite{yang2018hotpotqadatasetdiverseexplainable}, StrategyQA ~\cite{geva2021did}, MultiRC ~\cite{khashabi2018looking}), scientific reasoning (ARC ~\cite{clark2018think}), logical reasoning (FOLIO, P-FOLIO ~\cite{han2024pfolio, han2022folio}) constraint satisfaction puzzles (Game of 24 ~\cite{yao2023tree}), everyday common sense (MUSR ~\cite{sprague2023musr}), and challenging reasoning tasks (BBH ~\cite{suzgun2022bbh}). Several of these benchmarks, particularly HotpotQA, ALFWorlds, and Game of 24, have been specifically adapted for evaluating agent-based approaches like ReAct, where planning and calling the tools proposed by the agent are interleaved in interactive problem-solving settings.
 
Recent work has developed more specialized frameworks targeting LLM planning capabilities. ToolEmu ~\cite{ruan2024identifyingriskslmagents} introduces a simulator-based approach for evaluating tool-using agents, revealing that successful planning requires explicit state tracking and the ability to recover from errors. The MINT benchmark \cite{wang2023mint} evaluates planning in interactive environments, demonstrating that even advanced LLMs struggle with long-horizon tasks that require multiple steps. 

PlanBench ~\cite{valmeekam2023planbench} provides a comprehensive evaluation framework specifically designed to assess planning capabilities in LLM agents across diverse domains, revealing that current models excel at short-term tactical planning but struggle with strategic long-horizon planning. Complementing this, AutoPlanBench ~\cite{stein2023autoplanbench} focuses on evaluating planning in everyday scenarios, demonstrating that even SoTA LLM agents lag behind classical symbolic planners.

FlowBench \cite{xiao2024flowbench} evaluates workflow planning abilities, with a focus on expertise-intensive tasks. ACPBench ~\cite{kokel2024acpbench} focuses on evaluating LLMs on core reasoning skills. The Natural Plan benchmark ~\cite{zheng2024natural} is designed to evaluate how LLMs handle real-world planning tasks presented in natural language. SoTA LLM agents perform poorly on this benchmark, particularly as complexity increases.

These benchmarks highlight key abilities essential for effective agent planning: (1) task decomposition for breaking down complex problems, (2) state tracking and belief maintenance for accurate multi-step reasoning, (3) self-correction to detect and recover from errors, (4) causal understanding to predict action outcomes, and (5) meta-planning to refine planning strategies.

\subsection{Function Calling \& Tool Use} \label{app:tool}

The ability of LLMs to interact with external tools through function calling is fundamental for building intelligent agents capable of delivering real-time, contextually accurate responses \cite{qin2023toolllm, tang2023toolalpaca}. 
Early works utilized targeted tools, such as retrieval in approaches by augmented language models with retrieval capabilities \cite{lewis2020retrieval, gao2023retrieval, nakano2021webgpt}.
Later developments included more general-purpose tools, exemplified by ToolFormer \cite{schick2023toolformer}, Chameleon \cite{lu2023chameleon}, and MRKL \cite{karpas2022mrkl}.

Function calling involves several 
sub-tasks that work together seamlessly. Intent recognition identifies when a function is needed based on user requests. Function selection determines the most appropriate tool for the task. Parameter-value-pair mapping extracts relevant arguments from the conversation and assigns them to function parameters. Function execution invokes the selected function with those parameters to interact with external systems. Finally, response generation processes the function output and incorporates it into the LLM's reply to the user. This integrated process ensures accurate and efficient function calling within the LLM's workflow.

Early evaluation efforts offered approaches to evaluate the above sub-tasks while focusing on relatively simple, one-step interactions with explicitly provided parameters.  Benchmarks such as ToolAlpaca \cite{tang2023toolalpaca}, APIBench \cite{patil2025gorilla}, ToolBench \cite{qin2023toolllm}, and the Berkeley Function Calling Leaderboard v1 (BFCL) \cite{berkeley-function-calling-leaderboard} exemplify this phase, employing synthetic datasets and rule-based matching (e.g., via Abstract Syntax Trees) to establish baseline metrics like pass rates and structural accuracy. However, these methods were limited in capturing the complexities of real-world scenarios, which might include multistep conversations, parameters that are not explicitly mentioned in the conversation, and tools with complex input structures and long, intricate outputs.
BFCL v2 and v3 address these gaps by adding organizational tools and integrated multi-turn, multi-step evaluation logic, offering a more realistic simulation and highlighting the need for continuous state management.


Complementing this evolution, several benchmarks have broadened the evaluation landscape. For example, ToolSandbox \cite{lu2024toolsandbox} differs from previous benchmarks by incorporating stateful tool execution, implicit state dependencies, on-policy conversational evaluation with a built-in user simulator, and dynamic evaluation strategies for intermediate and final milestones across arbitrary trajectories. Seal-Tools \cite{wu2024seal} adopts a self-instruct \citep{wang2022self} methodology to generate nested tool calls, effectively modeling layered and interdependent interactions. In parallel, API-Bank \cite{li2023apibankcomprehensivebenchmarktoolaugmented} emphasizes realistic API engagements by utilizing dialogue-based evaluations and extensive training datasets. Frameworks like NexusRaven \cite{nexusraven} further enrich this landscape by focusing on generalized tool-use scenarios that mirror the diverse challenges encountered in practice. API-Blend \citep{basu2024api} suggested a comprehensive approach focusing on identifying, curating, and transforming existing datasets into a large corpus for training and systematic testing of tool-augmented LLMs. API-Blend mimics real-world scenarios involving API tasks such as API/tool detection, slot filling, and sequencing of detected APIs, providing utility for both training and benchmarking purposes. RestBench \citep{song2023restgpt} facilitates exploration of utilizing multiple APIs to address complex real-world user instructions. APIGen \cite{liu2024apigen} provides a comprehensive automated data generation pipeline that synthesizes high-quality function-calling datasets verified through hierarchical stages. StableToolBench \cite{guo2024stabletoolbench} addresses the challenges of function-calling evaluation by introducing a virtual API server with caching and simulators to alleviate API status changes.

Addressing the inherent complexity of multi-step interactions, ComplexFuncBench \cite{zhong2025complexfuncbench} was specifically designed to assess scenarios requiring implicit parameter inference, adherence to user-defined constraints, and efficient long-context processing. NESTFUL \cite{basu2024nestful}  focuses on adding complexity by evaluating LLMs on nested sequences of API calls where outputs from one call serve as inputs to subsequent calls. 

\subsection{Self-Reflection} \label{app:self_reflect}

An emerging line of research focuses on whether agents can self-reflect and improve their reasoning through interactive feedback, thereby reducing errors in multi-step interactions. This requires the model to understand the feedback and dynamically update its beliefs to carry out adjusted actions or reasoning steps over extensive trajectories.

Early efforts to gauge LLM agent self-reflection were often indirect, repurposing existing reasoning or planning tasks, such as AGIEval \citep{zhong2023agievalhumancentricbenchmarkevaluating}, MedMCQA \citep{pal2022medmcqalargescalemultisubject}, ALFWorld \citep{shridhar2021alfworldaligningtextembodied}, MiniWoB++ \citep{liu2018reinforcementlearningwebinterfaces}, etc., into multi-turn feedback loops, to see if models could recognize or correct their own errors given external feedback in confined settings~\citep{renze2024selfreflectionllmagentseffects,huang2024largelanguagemodelsselfcorrect,Shinn2023ReflexionLA,you-etal-2024-llm,sun2023adaplanneradaptiveplanningfeedback,liu2025selfreflectionmakeslargelanguage}. Improvement was typically measured by determining if the final answer was corrected, providing only a coarse evaluation and potentially ill-defined measurement, as observed improvements may depend on specific prompting techniques lacking proper standardization \citep{huang2024largelanguagemodelsselfcorrect,liu2025selfreflectionmakeslargelanguage}.

LLF-Bench \citep{cheng2023llfbenchbenchmarkinteractivelearning} was introduced to standardize benchmarks for interactive self-reflection. It includes diverse decision-making tasks and treats task instructions as part of the environment. To reduce overfitting, it allows randomization of task instruction descriptions and agent feedback.

Similarly, LLM-Evolve \citep{you-etal-2024-llm} was introduced to evaluate LLM agents' self-reflection capabilities on standard benchmarks such as MMLU \citep{hendrycks2020measuring}. This approach evaluates agents based on past experiences by collecting previous queries with feedback and extracting them as in-context demonstrations. To provide more granular insights into different feedback types, \citep{pan2025benchmarkstalkreevaluatingcode} focused specifically on coding agents, extending existing coding benchmarks like APPS \citep{hendrycks2021measuringcodingchallengecompetence} and LiveCodeBench \citep{jain2024livecodebenchholisticcontaminationfree} to interactive settings.

From a cognitive science perspective, Reflection-Bench \citep{li2024reflectionbenchprobingaiintelligence} was designed to assess LLMs' cognitive reflection capabilities, breaking down reflection into components like perception of new information, memory usage, belief updating following surprise, decision-making adjustments, counterfactual reasoning, and meta-reflection.

\subsection{Memory} \label{app:memory}

Memory mechanisms in LLM-based agents improve their handling of long contexts and information retrieval, overcoming static knowledge limits and supporting reasoning and planning in dynamic scenarios~\cite{park2023generative}. Unlike tool use, which connects agents to external resources, memory ensures context retention for extended interactions like processing documents or maintaining conversations. Agents rely on short-term memory for real-time responses and long-term memory for deeper understanding and applying knowledge over time. Together, these memory systems allow LLM-based agents to adapt, learn, and make well-informed decisions in tasks requiring persistent information access.

One prominent line of research focuses on addressing the challenge of limited context lengths in LLMs by incorporating memory mechanisms to enhance reasoning and retrieval across extended contexts and conversations.  Recent works, such as ReadAgent~\cite{lee2024humaninspiredreadingagentgist}, MemGPT~\cite{packer2024memgptllmsoperatingsystems}, and A-MEM~\cite{xu2025amemagenticmemoryllm}, investigate these methods and evaluate their efficacy through reasoning and retrieval metrics.

Specifically, ReadAgent structures reading by grouping content, condensing episodes into memories, and retrieving passages, with effectiveness shown on datasets like QUALITY~\cite{pang2021quality}, NarrativeQA~\cite{kovcisky2018narrativeqa}, and QMSum~\cite{zhong2021qmsum}.

Similarly, A-MEM introduces an advanced memory architecture evaluated using the LoCoMo benchmark \cite{maharana2024evaluating}, while MemGPT manages a tiered memory system tested on NaturalQuestions-Open  \cite{liu2024lost} and multi-session chat datasets~\cite{xu2021beyond}.

For episodic memory evaluation, \cite{huet2025episodicmemoriesgenerationevaluation} proposes a specialized benchmark to assess how LLMs generate and manage memories that capture specific events with contextual details. This benchmark utilizes synthetically created book chapters and events with LLMs-based judge evaluation metrics to measure accuracy and relevance. 
StreamBench~\cite{wu2024streambench} represents a more challenging setting, evaluating how agents leverage external memory components—including the memory of previous interactions and external feedback—to continuously improve performance over time, with quality and efficiency assessed across diverse datasets including text-to-SQL tasks (e.g., Spider~\cite{yu2018spider}), ToolBench~\cite{xu2023tool}, and HotpotQA~\citep{yang2018hotpotqadatasetdiverseexplainable}.

Beyond context length optimization, memory mechanisms also enhance real-time decision-making and learning in agent settings, focusing on action optimization~\cite{Liu2024FromLT, Shinn2023ReflexionLA, wang2024karmaaugmentingembodiedai}. For example, Reflexion~\citep{Shinn2023ReflexionLA} tracks success rate on tasks like HotPotQA~\citep{yang2018hotpotqadatasetdiverseexplainable} and ALFWorld~\citep{shridhar2021alfworldaligningtextembodied}, while RAISE~\citep{Liu2024FromLT} enhances the ReAct framework with a two-part memory system evaluated through human judgment on quality metrics and efficiency. Similarly, KARMA~\citep{wang2024karmaaugmentingembodiedai} tests memory in household tasks using metrics such as success rate, retrieval accuracy, and memory hit rate, demonstrating how memory mechanisms significantly improve agent performance across diverse domains requiring complex reasoning and persistent information retention. LTMbenchmark~\citep{castillo-bolado2024beyond} evaluates conversational agents through extended, multi-task interactions with frequent context switching to test long-term memory and information integration capabilities. The results demonstrate that while LLMs generally perform well in single-task scenarios, they struggle with interleaved tasks, and interestingly, short-context LLMs equipped with long-term memory systems can match or exceed the performance of models with larger context windows.

\section{Application-Specific Agents Evaluation}

\subsection{Scientific Agents}

\paragraph{Scientific frameworks}
Complementing task-level benchmarks, unified evaluation frameworks are introduced to assess agents across sequential, end-to-end research workflows. These efforts range from gym-style workflow simulators ~\cite{nathani2025mlgym} and virtual discovery environments~\cite{jansen2024discoveryworldvirtualenvironmentdeveloping} to domain-focused suites like LAB-Bench~\cite{laurent2024labbenchmeasuringcapabilitieslanguage} for biologically orientated evaluation. Such frameworks enable holistic measurement of experimental design, iterative hypothesis refinement, and capabilities central to autonomous, innovative scientific discovery.

\paragraph{Deep Research}
Recently, research agents at the intersection of search, web, and scientific agents have become common, due to a wide commercial offering. Such agents retrieve and synthesize information before returning comprehensive citation-backed answers. This process poses a difficult 
evaluation task with multi-step, multi-faceted, time-insensitive consideration.
\citet{patel2025deepscholarbenchlivebenchmarkautomated} proposed evaluating those agents based on three core functions: retrieval quality, knowledge synthesis, and verifiability. \citet{gou2025mind2web2evaluatingagentic} utilized agent-as-a-judge for evaluation, and shows agents achieve $50-70\%$ of human performance. This is an active research field with many works \cite{wei2025browsecompsimplechallengingbenchmark,chen2025browsecomp, xu2025researcherbench, bosse2025deep, du2025deepresearch}.

\section{Generalist Agent Evaluation}\label{app:general}

\paragraph{Realistic Workplaces Benchmarks}
Another venue of generalist agent evaluation focuses on realistic workplaces with different roles and personas. 
TheAgentCompany~\cite{xu2024theagentcompany} simulates a software company where agents must browse internal sites, write code, and collaborate. CRMArena~\cite{huang2025crmarena} replicates a Customer Relationship Management environment, requiring agents to use UI and APIs, follow domain policies, and integrate diverse data to complete enterprise-level tasks.

\section{Benchmark Recommendations}
\label{app:recommendations}

Navigating the extensive landscape of agent evaluation is a prerequisite for effective benchmarking. In this section, we distill our survey findings into actionable recommendations for practitioners and researchers. Our selection criteria prioritize community adoption, active maintenance status, and the reporting standards observed in recent literature from leading LLM and agent developers.

\subsection{Web Agents}
The choice of benchmark in this domain depends heavily on the environment's dynamicity and the agent's modality.
\begin{itemize}
    \item \textbf{Dynamic Interaction:} \textbf{WebArena} remains the leading option for agents operating in dynamic environments. We note that as of April 19, 2026, top performance has reached 74.3\%, based on a submission from February 2026, suggesting there is still room for improvement.
    \item \textbf{Static Evaluation:} For offline evaluation using cached traces, \textbf{Mind2Web} remains the standard.
    \item \textbf{Online \& Multimodal:} For agents heavily reliant on visual signals, \textbf{WebVoyager} is the recommended choice. Researchers seeking updated, reactive environments should also consider the new but promising \textbf{Mind2Web-Live} and \textbf{Online-Mind2Web}.
\end{itemize}

\subsection{Software Engineering (SWE) Agents}
\textbf{SWE-bench Verified} continues to serve as the dominant gold standard for evaluating coding agents. However, with top performances now reaching approximately 80\%, suggesting it is close to being saturated.
\begin{itemize}
    \item \textbf{Higher Difficulty:} We recommend \textbf{SWE-bench-pro} (Scale) as a more rigorous alternative, where current state-of-the-art (SOTA) performance hovers around 46\%.
    \item \textbf{Specialized Contexts:} \textbf{SWE-Lancer} is valuable for evaluating freelance-style task completion, while \textbf{Terminal Bench} is essential for agents specializing in command-line interface (CLI) interactions.
    \item \textbf{Multi options:} Multilingual and multi-modal focused benchmarks are mentioned in the paper. 
\end{itemize}

\subsection{Scientific Agents}
Evaluation in scientific domains is highly task-dependent. In the main paper, we try to mention the best options for each scientific task. 

\subsection{Conversational Agents}
For agents requiring robust user simulation and tool usage in dialogue, \textbf{$\tau$-bench} is the community standard. While current models achieve high success rates on this benchmark, it remains the most common option, and no widely adopted alternative has yet emerged to displace it.

\subsection{Generalist Agent Evaluation}
For general-purpose agents, the recommendation varies by the task focus and the agent's interaction interface:
\begin{itemize}
    \item \textbf{Reasoning Focus:} \textbf{GAIA} remains the primary recommendation for testing general logical reasoning and tool selection.
    \item \textbf{Tools Interface:} \textbf{AppWorld} is the preferred environment for agents that interact via coding or structured tool calling.
    \item \textbf{GUI Interface:} For agents interacting via User Interface, \textbf{OS-World} is the standard.
    \item \textbf{Holistic Evaluation:} For cross-benchmark assessment, we recommend starting with the \textbf{HAL} (Holistic Agent Leaderboard) framework to standardize comparisons across these disparate domains.
\end{itemize}

\newpage

%% file: figures/layout_figs.tex
\definecolor{paired-light-blue}{RGB}{198, 219, 239}
\definecolor{paired-dark-blue}{RGB}{49, 130, 188}
\definecolor{paired-light-orange}{RGB}{251, 208, 162}
\definecolor{paired-dark-orange}{RGB}{230, 85, 12}
\definecolor{paired-light-green}{RGB}{199, 233, 193}
\definecolor{paired-dark-green}{RGB}{49, 163, 83}
\definecolor{paired-light-purple}{RGB}{218, 218, 235}
\definecolor{paired-dark-purple}{RGB}{117, 107, 176}
\definecolor{paired-light-gray}{RGB}{217, 217, 217}
\definecolor{paired-dark-gray}{RGB}{99, 99, 99}
\definecolor{paired-light-pink}{RGB}{222, 158, 214}
\definecolor{paired-dark-pink}{RGB}{123, 65, 115}
\definecolor{paired-light-red}{RGB}{231, 150, 156}
\definecolor{paired-dark-red}{RGB}{131, 60, 56}
\definecolor{paired-light-yellow}{RGB}{231, 204, 149}
\definecolor{paired-dark-yellow}{RGB}{141, 109, 49}
\definecolor{light-green}{RGB}{118, 207, 180}
\definecolor{raspberry}{RGB}{228, 24, 99}

\tikzset{%
    root/.style =          {align=center,text width=3cm,rounded corners=3pt, line width=0.5mm, fill=paired-light-gray!50,draw=paired-dark-gray!90},
    data_section/.style =  {align=center,text width=4cm,rounded corners=3pt, fill=paired-light-blue!50,draw=paired-dark-blue!80,line width=0.4mm},
    model_section/.style = {align=center,text width=4cm,rounded corners=3pt, fill=paired-light-orange!50,draw=paired-dark-orange!80,line width=0.4mm},
    training_section/.style = {align=center,text width=4cm,rounded corners=3pt, fill=paired-light-green!50,draw=paired-dark-green!80, line width=0.4mm},
    inference_section/.style = {align=center,text width=4cm,rounded corners=3pt, fill=paired-light-red!35,draw=paired-light-red!90, line width=0.4mm},
    discussion_section/.style = {align=center,text width=4cm,rounded corners=3pt, fill=paired-light-purple!35,draw=paired-dark-purple!90, line width=0.4mm},
    subsection/.style =    {align=center,text width=3.5cm,rounded corners=3pt}, 
}

\begin{figure*}[!htb]
    \centering
    \resizebox{1\textwidth}{!}{
    \begin{forest}
        for tree={
            forked edges,
            grow'=0,
            draw,
            rounded corners,
            node options={align=center},
            text width=4cm,
            s sep=6pt,
            calign=child edge,
            calign child=(n_children()+1)/2,
            l sep=12pt,
        },
        [Agent Evaluation, root, , calign child=2,
            [Agent Capabilities Evaluation (\S\ref{sec:capabilities}), data_section
                [Planning and Multi-Step Reasoning (\S\ref{sec:plan}), data_section 
                    [ 
                    \textit{HotpotQA} ~\cite{yang2018hotpotqadatasetdiverseexplainable};
                    \textit{Game of 24} ~\cite{yao2023tree};
                    \textit{MINT} ~\cite{wang2023mint};
                    \textit{PlanBench} ~\cite{valmeekam2023planbench};
                    \textit{FlowBench} ~\cite{xiao2024flowbench};
                    \textit{MultiRC} ~\cite{khashabi2018looking};
                    \textit{MUSR} ~\cite{sprague2023musr};
                    \textit{ToolEmu} ~\cite{ruan2024identifyingriskslmagents};
                    \textit{AutoPlanBench} ~\cite{stein2023autoplanbench};
                    \textit{ACPBench} ~\cite{kokel2024acpbench};
                    \textit{Natural Plan} ~\cite{zheng2024natural}
                    ,data_section, text width=12cm
                    ] 
                ]
                [Function Calling \& Tool Use (\S\ref{sec:tool}), data_section
                    [\textit{BFCL} \cite{berkeley-function-calling-leaderboard};
                    \textit{ToolBench} \cite{qin2023toolllm};
                    \textit{ToolAlpaca} \cite{tang2023toolalpaca};
                    \textit{APIBench} \cite{peng2021revisitingbenchmarkingexploringapi};
                    \textit{API-Bank} \cite{li2023apibankcomprehensivebenchmarktoolaugmented};
                    \textit{NexusRaven} \cite{nexusraven};
                    \textit{Seal-Tools} \cite{wu2024seal};
                    \textit{ComplexFuncBench} \cite{zhong2025complexfuncbench};
                    \textit{ToolSandbox} \cite{lu2024toolsandbox};
                    \textit{RestBench} \citep{song2023restgpt};
                    \textit{APIGen} \cite{liu2024apigen};
                    \textit{StableToolBench} \cite{guo2024stabletoolbench};
                    \textit{NESTFUL} \cite{basu2024nestful};
                    \textit{MCP-Universe} \cite{luo2025mcpuniversebenchmarkinglargelanguage},
                    data_section, text width=12cm
                    ] 
                ]
                [Self-Reflection (\S\ref{sec:self_reflect}), data_section
                    [\textit{LLF-Bench} \citep{cheng2023llfbenchbenchmarkinteractivelearning};
                    \textit{LLM-Evolve} \citep{you-etal-2024-llm};
                    \textit{Reflection-Bench}
                    \citep{li2024reflectionbenchprobingaiintelligence},
                    data_section, text width=12cm
                    ] 
                ]
                [Memory (\S\ref{sec:memory}), data_section
                    [\textit{NarrativeQA} ~\cite{kovcisky2018narrativeqa};
                    \textit{QMSum} ~\cite{zhong2021qmsum};
                    \textit{QUALITY} ~\cite{pang2021quality};
                    \textit{RAISE}\citep{Liu2024FromLT};
                    \textit{ReadAgent} ~\cite{lee2024humaninspiredreadingagentgist};
                    \textit{MemGPT} ~\cite{packer2024memgptllmsoperatingsystems};
                    \textit{LoCoMo} ~\cite{maharana2024evaluating};
                    \textit{A-MEM} ~\cite{xu2025amemagenticmemoryllm};
                    \textit{StreamBench} ~\cite{wu2024streambench};
                    \textit{LTMbenchmark}
                    ~\citep{castillo-bolado2024beyond},
                    data_section, text width=12cm
                    ] 
                ]
            ]
            [Application-Specific Agent Evaluation (\S\ref{sec:use_case}), model_section
                [Web Agents (\S\ref{sec:web}), model_section 
                    [ \textit{MiniWob} \citep{pmlr-v70-shi17a};
                    \textit{MiniWoB++} \citep{liu2018reinforcementlearningwebinterfaces};
                    \textit{WebShop} \cite{yao2022webshop};
                    \textit{Mind2web} \cite{deng2023mind2web};
                    \textit{WebVoyager} \cite{he2024webvoyager};
                    \textit{WebLinX} \cite{lu2024weblinx};
                    \textit{WebArena} \citep{zhou2023webarena};
                    \textit{VisualWebArena} \citep{koh2024visualwebarena};
                    \textit{MMInA} \cite{zhang2024mmina};
                    \textit{AssistantBench} \cite{yoran2024assistantbench};
                    \textit{WebCanvas} \cite{pan2024webcanvas};
                    \textit{ST-WebAgentBench} \cite{levy2024st};
                    \textit{WorkArena} \citep{drouin2024workarena};
                    \textit{WorkArena++} \citep{boisvert2025workarena++};
                    \textit{WindowsAgentArena} \cite{bonatti2025windows};
                    \textit{SafeArena} \cite{tur2025safearena},
                    model_section, text width=12cm] 
                ]
                [Software Engineering \\ Agents (\S\ref{sec:se}), model_section
                    [\textit{HumanEval} \cite{Chen2021EvaluatingLL};
                    \textit{SWE-bench}~\cite{Jimenez2023SWEbenchCL};
                    \textit{SWE-bench Verified}~\cite{swebenchverified2024};
                    \textit{SWE-bench Lite}~\cite{swebenchlite2024};
                    \textit{SWE-bench+} \cite{Aleithan2024SWEBenchEC};
                    \textit{SWE-bench Multimodal} \cite{Yang2024SWEbenchMD};
                    \textit{TDD-Bench Verified} \cite{ahmed2024tdd};
                    \textit{SWT-Bench} \cite{mundler2024swt};
                    \textit{ITBench}~\cite{jha2025itbench};
                    \textit{SWELancer} \cite{miserendino2025swe},
                    model_section, text width=12cm] 
                ]
                [Scientific Agents (\S\ref{sec:sa}), model_section
                    [
                    \textit{ScienceQA} \cite{lu2022learnexplainmultimodalreasoning};
                    \textit{QASPER} \cite{dasigi2021datasetinformationseekingquestionsanswers};
                    \textit{SciRIFF-Eval} \cite{wadden2024sciriff}; 
                    \textit{MS\textsuperscript{2}} \cite{deyoung2021ms2multidocumentsummarizationmedical};
                    \textit{SciDQA} \cite{singh-etal-2024-scidqa};                     
                    \textit{ScienceWorld}\cite{wang-etal-2022-scienceworld};
                    \textit{SUPER} \citep{bogin2024superevaluatingagentssetting};
                    \textit{Ideation}~\cite{si2025can};
                    \textit{AAAR-1.0}~\cite{lou2025aaar};
                    \textit{ScienceAgentBench} \citep{chen2025scienceagentbench};
                    \textit{CORE-Bench} \citep{siegel2024corebenchfosteringcredibilitypublished};
                    PaperBench~\cite{starace2025paperbenchevaluatingaisability};
                    \textit{SciCode} \citep{tian2024scicode};
                    \textit{MLGym-Bench} \citep{nathani2025mlgym};
                    \textit{DiscoveryWorld} \citep{jansen2024discoveryworldvirtualenvironmentdeveloping};
                    \textit{LAB-Bench} \cite{laurent2024labbenchmeasuringcapabilitieslanguage},
                    model_section, text width=12cm] 
                ]
                [Conversational Agents (\S\ref{sec:conv}), model_section
                    [\textit{MultiWOZ }\cite{budzianowski-etal-2018-multiwoz};
                    \textit{SMCalFlow} \cite{andreas-etal-2020-task};
                    \textit{ABCD} \cite{chen-etal-2021-action};
                    \textit{ALMITA} \cite{arcadinho-etal-2024-automated};
                    \textit{$\tau$-Bench} \cite{yao2024taubenchbenchmarktoolagentuserinteraction};
                    \textit{IntellAgent} \cite{levi2025intellagentmultiagentframeworkevaluating};
                    \textit{LTM} \cite{castillo-bolado2024beyond}
                    , model_section, text width=12cm] 
                ]
            ]
            [Generalist Agents Evaluation (\S\ref{sec:general}), training_section
                [, training_section
                [\textit{GAIA} \cite{mialon2023gaia};
                \textit{AgentBench}~\cite{liu2023agentbench};
                \textit{Galileo’s Agent Leaderboard}~\cite{agent-leaderboard};
                \textit{OSWorld}~\cite{xie2024osworld};
                \textit{AppWorld}~\cite{trivedi-etal-2024-appworld};
                \textit{OmniACT}~\cite{omniact};
                \textit{TheAgentCompany}~\cite{xu2024theagentcompany};
                \textit{CRMArena}~\cite{huang2025crmarena};
                \textit{HAL}~\cite{hal},
                training_section, text width=12cm]
                ]
            ]
            [Frameworks for Agent Evaluation
            (\S\ref{sec:frameworks}), inference_section
                [Development Frameworks, inference_section
                [
                \textit{Databricks Mosaic AI} \citep{mosaic_ai_agent_evaluation};
                 \textit{Galileo Agentic} \citep{galileo_agentic_evaluations};
                \textit{Vertex AI Gen AI} \citep{vertex_ai_gen_ai_evaluation};
                \textit{LangSmith} \citep{langsmith};
                \textit{Langfuse} \citep{langfuse};
                \textit{Patronus AI} \citep{patronus_ai};
                \textit{LangChain AgentEvals} \cite{agentevals},
                inference_section, text width=12cm]
                ] 
                [Gym-like Environments, inference_section
                [
                \textit{MLGym}\citep{nathani2025mlgym};
                \textit{BrowserGym}\citep{chezelles2024browsergym};
                \textit{SWE-Gym}\citep{pan2024training},
                inference_section, text width=12cm]
                ] 
            ]
        ]
    \end{forest}
    }
    \caption{Overview of the paper.} 
    \label{fig:agent-evaluation-typology_full}
\end{figure*}